\documentclass[11pt]{article}
\usepackage{coling2016}
\usepackage{times}
\usepackage{latexsym}
\usepackage{amsmath}
\usepackage{amssymb}
\usepackage{graphicx}
\graphicspath{ {images/} }
\usepackage{verbatim}
\usepackage{fancyvrb}
\usepackage{multirow}
\usepackage{multicol}
\usepackage{hhline}
\usepackage{url}
\usepackage{color}
\usepackage[labelfont=bf]{caption}
\usepackage{tikz}
\usepackage{standalone}
\usepackage{colortbl}
\usepackage{tabularx}
\usepackage{xcolor}
\usepackage{pgf}
\usepackage{collcell}
\usepackage{framed}

\newlength{\bibitemsep}
\newlength{\bibparskip}
\let\oldthebibliography\thebibliography
\renewcommand\thebibliography[1]{%
  \oldthebibliography{#1}%
  \setlength{\parskip}{\bibitemsep}%
  \setlength{\itemsep}{\bibparskip}%
}
	
\definecolor{brinkpink}{rgb}{0.98, 0.38, 0.5}

\title{CogALex-V Shared Task:\\LexNET - Integrated Path-based and Distributional Method\\for the Identification of Semantic Relations}

\author{Vered Shwartz ~~~~~~~~~~~~~~~~~~~~~~~~~~~~~~~~~~~~~~~ Ido Dagan\\
   Computer Science Department, Bar-Ilan University, Ramat-Gan, Israel\\
   {\tt vered1986@gmail.com	~~~~~~	\tt dagan@cs.biu.ac.il }}

\date{}

\begin{document}

\maketitle

\begin{abstract}
We present a submission to the CogALex 2016 shared task on the corpus-based identification of semantic relations, using \texttt{LexNET} \cite{shwartz2016roles}, an integrated path-based and distributional method for semantic relation classification. 
The reported results in the shared task bring this submission to the third place on subtask 1 (word relatedness), and the first place on subtask 2 (semantic relation classification), demonstrating the utility of integrating the complementary path-based and distributional information sources in recognizing concrete semantic relations.
Combined with a common similarity measure, \texttt{LexNET} performs fairly good on the word relatedness task (subtask 1).
The relatively low performance of \texttt{LexNET} and all other systems on subtask 2, however, confirms the difficulty of the semantic relation classification task, and stresses the need to develop additional methods for this task. 
\end{abstract}

\section{Introduction} 

\blfootnote{
     
    
     \hspace{-0.65cm}  
     This work is licenced under a Creative Commons 
     Attribution 4.0 International License.
     License details:\\
     \url{http://creativecommons.org/licenses/by/4.0/}
}

Discovering whether words are semantically related and identifying the specific semantic relation that holds between them is a key component in many NLP applications, such as question answering and recognizing textual entailment \cite{dagan2013recognizing}. Automated methods for semantic relation identification are commonly corpus-based, and mainly rely on the distributional representation of each word.

The CogALex shared task on the corpus-based identification of semantic relations consists of two subtasks. In the first task, the system needs to identify for a word pair whether the words are semantically related or not (e.g. \emph{True:(dog, cat), False:(dog, fruit)}). In the second task, the goal is to determine the specific semantic relation that holds for a given pair, if any (\emph{PART\_OF:(tail, cat), HYPER:(cat, animal)}).

In this paper we describe our approach and system setup for the shared task. We use \texttt{LexNET} \cite{shwartz2016roles}, an integrated path-based and distributional method for semantic relation classification. \texttt{LexNET} was the system with the overall best performance on subtask 2, and was ranked third on subtask 1, demonstrating the utility of integrating the complementary path-based and distributional information sources in recognizing semantic relatedness.\footnote{LexNET's code is available at \url{https://github.com/vered1986/LexNET}, and the shared task results are available at \url{https://sites.google.com/site/cogalex2016/home/shared-task/results}}

To aid in recognizing whether a pair of words are related at all (subtask 1), we combine \texttt{LexNET} with a common similarity measure (cosine similarity), achieving fairly good performance, and a slight improvement upon using cosine similarity alone. Subtask 2, however, has shown to be extremely difficult, with \texttt{LexNET} and all other systems achieving relatively low $F_1$ scores. The conflict between the mediocre performance and the recent success of distributional methods on several other common datasets for semantic relation classification \cite{baroni2012entailment,weeds2014learning,roller2014inclusive} could be explained by the stricter evaluation setup in this subtask, which is supposed to demonstrate more closely real-world application settings. The difficulty of the semantic relation classification task emphasizes the need to develop better methods for this task.

\section{Background}
\label{sec:background}

\subsection{Word Relatedness}
\label{sec:bg_relatedness}

Recognizing word relatedness is typically addressed by distributional methods. To determine to what extent two terms $x$ and $y$ are related, a vector similarity or distance measure is applied to their distributional representations: $sim(\vec{v}_{w_x}, \vec{v}_{w_y})$. This is a straightforward application of the distributional hypothesis \cite{harris1954distributional}, according to which related words occur in similar contexts, hence have similar vector representations.  

Most commonly, vector cosine is adopted as a similarity measure \cite{turney2010frequency}. Many other measures exist, including but not limited to Euclidean distance, KL divergence \cite{cover2012elements}, Jaccard's coefficient \cite{salton1986introduction}, and more recently neighbor rank \cite{hare2009activating,lapesa2013evaluating} and APSyn \cite{santus2016testing}.\footnote{See \newcite{lee1999measures} for an extensive list of such measures.} To turn this task into a binary classification task, where $x$ and $y$ are classified as either related or not, one can set a threshold to separate similarity scores of related and unrelated word pairs. 

\subsection{Semantic Relation Classification}
\label{sec:bg_sem_rel}

Recognizing lexical semantic relations between words is valuable for many NLP applications, such as ontology learning, question answering, and recognizing textual entailment. Most corpus-based methods classify the relation between a pair of words $x$ and $y$ based on the distributional representation of each word \cite{baroni2012entailment,roller2014inclusive,fu2014learning,weeds2014learning}. Earlier methods utilized the dependency paths that connect the joint occurrences of $x$ and $y$ in the corpus as a cue to the relation between the words \cite{hearst1992automatic,snow2004learning,nakashole2012patty}. Recently, \newcite{shwartz2016roles} presented \texttt{LexNET}, an extension of \texttt{HypeNET} \cite{shwartz2016improving}. This method integrates both path-based and distributional information for semantic relation classification, which outperformed approaches that rely on a single information source, on several common datasets \cite{baroni2011we,necsulescu2015reading,santus2015evalution,santus2016nine}.

\section{System Description}
\label{sec:system}

In \texttt{LexNET}, a word-pair $(x, y)$ is represented as a feature vector, consisting of a concatenation of distributional and path-based features: $\vec{v}_{xy} = [\vec{v}_{w_x}, \vec{v}_{paths(x,y)}, \vec{v}_{w_y}]$, where $\vec{v}_{w_x}$ and $\vec{v}_{w_y}$ are $x$ and $y$'s word embeddings, providing their distributional representation, and $\vec{v}_{paths(x,y)}$ is the average embedding vector of all the dependency paths that connect $x$ and $y$ in the corpus. Dependency paths are embedded using a LSTM \cite{hochreiter1997long}, as described in \newcite{shwartz2016improving}. This vector is then fed into a neural network that outputs the class distribution $\vec{c}$, and then the pair is classified to the relation with the highest score $r$:

\vspace{-20pt}
\begin{subequations}
\begin{align}
    & \vec{c} = \operatorname{softmax}(\operatorname{MLP}(\vec{v}_{xy}))\\
		& r = \operatorname{argmax}_i \vec{c}[i]
\end{align}
\vspace{-12pt}
\end{subequations}

\noindent MLP stands for Multi Layer Perceptron, and could be computed with or without a hidden layer (equations \ref{eq:hidden} and \ref{eq:no-hidden}, respectively): 

\vspace{-20pt}
\begin{subequations}
\label{eq:hidden}
\begin{align}
    & \vec{h} = \operatorname{tanh}(W_1 \cdot \vec{v}_{xy} + b_1)\\
		& \operatorname{MLP}(\vec{v}_{xy}) = W_2 \cdot \vec{h} + b_2
\end{align}
\vspace{-20pt}
\end{subequations}

\vspace{-5pt}
\begin{equation}
\label{eq:no-hidden}
\operatorname{MLP}(\vec{v}_{xy}) = W_1 \cdot \vec{v}_{xy} + b_1
\end{equation}

\noindent where $W_i$ and $b_i$ are the network parameters and $\vec{h}$ is the hidden layer.

\subsection{A Note About Word Relatedness}
\label{sec:note}

While path-based approaches have been commonly used for semantic relation classification \cite{hearst1992automatic,snow2004learning,nakashole2012patty,necsulescu2015reading}, they have never been used for word relatedness, which is considered a ``classical'' task for distributional methods. We argue that path-based information can improve performance of word relatedness tasks as well (see Section~\ref{sec:subtask1}). We train \texttt{LexNET} to distinguish between two classes: \textsc{related} and \textsc{unrelated}, and combine it with the common cosine similarity measure to tackle subtask 1.

\section{Experimental Settings}
\label{sec:experimental}

\begin{table*}[!t]
\center
\small
\hspace*{-10pt}
\begin{tabular}{ c | c | c | c | c | c | c | }
    \hhline{~------}
		& \textbf{Method} & \textbf{Hyper-parameters} & \textbf{Corpus size} & \textbf{P} & \textbf{R} & \boldmath $F_1$ \\ \hline
		\multicolumn{1}{|c|}{\multirow{3}{*}{\textbf{Subtask 1}}} & \texttt{Cos} & word2vec, $t$: 0.3 & 100B & 0.759 & 0.795 & 0.776 \\ \hhline{~------}
		\multicolumn{1}{|c|}{} & \texttt{LexNET} & hidden layers: 0, dropout: 0.0, epochs: 3 & 6B & 0.780 & 0.561 & 0.652 \\ \hhline{~------}
		\multicolumn{1}{|c|}{} & \texttt{LexNET}+\texttt{Cos} & word2vec, $w_L=0.3,w_C=0.7,t=0.29$ & $\sim$100B & \textbf{0.814} & \textbf{0.854} & \textbf{0.833} \\ \hline
		\multicolumn{1}{|c|}{\multirow{2}{*}{\textbf{Subtask 2}}} & \texttt{Dist} & dep-based, method: concat, classifier: SVM, $L_1$ & 3B & 0.611 & 0.598 & 0.600 \\ \hhline{~------}
		\multicolumn{1}{|c|}{} & \texttt{LexNET} & hidden layers: 0, dropout: 0.0, epochs: 5 & 6B & \textbf{0.658} & \textbf{0.646} & \textbf{0.642} \\ \hline
	\end{tabular}
	\caption{Performance scores on the validation set along with hyper-parameters and effective corpus size (\#tokens) used by each method. Subtask 2 results refer to the subset of related pairs, as detailed in \S~\ref{sec:subtask2}.}
	\label{tab:validation}
\end{table*}

The shared task organizers provided a dataset extracted from EVALution 1.0 \cite{santus2015evalution}, which was split into training and test sets. As instructed, we trained and tuned our method on the training set, and evaluated it once on the test set. To tune the hyper-parameters, we split the training set to 90\% train and 10\% validation sets. Since the dataset contains only 318 different words in the $x$ slot, we performed the split such that the train and the validation contain distinct $x$ words.\footnote{A random split yielded perfect results on the validation set, which were due to lexical memorization \cite{levy2015supervised}.}

\texttt{LexNET} has several tunable hyper-parameters. Similarly to \newcite{shwartz2016roles}, we used the English Wikipedia dump from May 2015 as an underlying corpus (3B tokens), and initialized the network's word embeddings with the 50-dimensional pre-trained GloVe word embeddings \cite{pennington2014glove}, trained on Wikipedia and Gigaword 5 (6B tokens). We fixed this hyper-parameter due to computational limitations with higher-dimensional embeddings. For each subtask, we tuned \texttt{LexNET}'s hyper-parameters on the validation set: the number of hidden layers (0 or 1), the number of training epochs, and the word dropout rate (see \newcite{shwartz2016improving} for technical details). Table~\ref{tab:validation} displays the best performing hyper-parameters in each subtask, along with the performance on the validation set, which is detailed below.

\subsection{Subtask 1: Word Relatedness}
\label{sec:subtask1}

We tuned \texttt{LexNET}'s hyper-parameters on the validation set, disregarding the similarity measure at this point, and then chose the model that performed best on the validation set and combined it with the similarity measure. 

We computed cosine similarity for each $(x, y)$ pair in the dataset: $\operatorname{cos}(\vec{v}_{w_x}, \vec{v}_{w_y}) = \frac{\vec{v}_{w_x} \cdot \vec{v}_{w_y}}{\|\vec{v}_{w_x}\| \cdot \|\vec{v}_{w_y}\|}$, and normalized it to the range $[0,1]$. We scored each $(x, y)$ pair by a combination of \texttt{LexNET}'s score for the \textsc{related} class and the cosine similarity score:

\begin{equation}
\label{eq:combined}
\operatorname{Rel}(x, y) = w_C \cdot \operatorname{cos}(\vec{v}_{w_x}, \vec{v}_{w_y}) + w_L \cdot \vec{c}[\textsc{related}]
\end{equation}

\noindent where $w_C, w_L$ are the weights assigned to cosine similarity and \texttt{LexNET}'s scores respectively, such that $w_C + w_L = 1$. We tuned the weights and a threshold $t$ using the validation set, and classified $(x, y)$ as related if $\operatorname{Rel(x, y)} \geq t$. The word vectors used to compute the cosine similarity scores were chosen among several available pre-trained embeddings.\footnote{word2vec (300 dimensions, SGNS, trained on GoogleNews, 100B tokens) \cite{mikolov2013distributed}, GloVe (50-300 dimensions, trained on Wikipedia and Gigaword 5, 6B tokens) \cite{pennington2014glove}, and dependency-based embeddings (300 dimensions, trained on Wikipedia, 3B tokens) \cite{levy2014dependency}.} For completeness we also report the performance of two baselines: cosine similarity ($w_C = 1$) and \texttt{LexNET} ($w_L = 1$, fixed $t = 0.5$).

\subsection{Subtask 2: Semantic Relation Classification}
\label{sec:subtask2}

The subtask's train set is highly imbalanced towards random instances (roughly 10 times more than any other relation), and training any supervised method leads to overfitting to the random class. We therefore trained the model only on the related classes (excluding \textsc{random} pairs), for which the classes are more balanced. During inference time, we used the model from subtask 1 to assign a relatedness score to each pair, $Rel(x, y)$, and computed the class distribution using the model from subtask 2, only for pairs that were related according to this score.

Finally, we applied a heuristic that if for a word pair $(x,y)$, the difference in scores between the top scoring classes is low ($< 0.2$), and the top class is \textsc{syn}, then it is only classified as \textsc{syn} if the number of paths between $x$~and~$y$ is smaller than 3. This is due to the fact that synonyms are hard to recognize with both distributional and path-based approaches \cite{shwartz2016roles}, but it is known that they do not tend to co-occur. 

To compare \texttt{LexNET}'s performance on the validation set with other methods' performances, we adapted the distributional baseline employed by \newcite{shwartz2016improving} and \newcite{shwartz2016roles}, where a classifier is trained on the combination of $x$ and $y$'s word embeddings. We experimented with several combination methods (concatenation \cite{baroni2012entailment}, difference \cite{fu2014learning,weeds2014learning}, and ASYM \cite{roller2014inclusive}), regularization factors, and pre-trained word embeddings \cite{mikolov2013distributed,pennington2014glove,levy2014dependency}. This time, we used cosine similarity (subtask 1) to separate related from unrelated pairs, and trained the classifier only to distinguish between the related classes. 
Similarly to subtask 1, we tune \texttt{LexNET} and the baseline's hyper-parameters on the validation set. The best performance is reported in Table~\ref{tab:validation}.

\begin{table*}[!t]
\center
\small
\begin{tabular}{ c | c | c | c | c | }
    \hhline{~----}
		& \textbf{Method} & \textbf{P} & \textbf{R} & \boldmath $F_1$ \\ \hline
		\multicolumn{1}{|c|}{\multirow{4}{*}{\textbf{Subtask 1}}} & Random Baseline & 0.283 & 0.503 & 0.362 \\ \hhline{~----}
		\multicolumn{1}{|c|}{} & Majority Baseline & 0.000 & 0.000 & 0.000 \\ \hhline{~----}
		\multicolumn{1}{|c|}{} & \texttt{Cos} & \textbf{0.841} & 0.672 & 0.747 \\ \hhline{~----}
		\multicolumn{1}{|c|}{} & \textbf{\texttt{LexNET}+\texttt{Cos}} & 0.754 & \textbf{0.777} & \textbf{0.765} \\ \hline
		\multicolumn{1}{|c|}{\multirow{4}{*}{\textbf{Subtask 2}}} & Random Baseline & 0.073 & 0.201 & 0.106 \\ \hhline{~----}
		\multicolumn{1}{|c|}{} & Majority Baseline & 0.000 & 0.000 & 0.000 \\ \hhline{~----}
		\multicolumn{1}{|c|}{} & \texttt{Dist} & 0.469 & 0.371 & 0.411 \\ \hhline{~----}
		\multicolumn{1}{|c|}{} & \textbf{\texttt{LexNET}} & \textbf{0.480} & \textbf{0.418} & \textbf{0.445} \\ \hline
	\end{tabular}
	\caption{Performance scores on the test set in each subtask, of the selected methods and the baselines.}
	\label{tab:test}
\end{table*} 

\section{Results and Analysis}
\label{sec:evaluation}

Table~\ref{tab:test} displays the performance of our methods and the baselines on the test set. In addition to the two baselines provided by the shared task organizers (majority and random), we report also the results of our baselines detailed in Section~\ref{sec:experimental}. The majority baseline classifies all the instances as \textsc{unrealted} (subtask 1) or \textsc{random} (subtask 2). Since these labels are excluded from the averaged $F_1$ computation, this baseline's performance metrics are all zero. 

\paragraph{Subtask 1: Word Relatedness}

\texttt{Cos} achieves fairly good performance ($F_1=0.747$), and \texttt{LexNET+Cos} slightly improves upon it. 
To better understand \texttt{LexNET}'s contribution, we examined pairs that were correctly classified by \texttt{LexNET+Cos} while being incorrectly classified by \texttt{Cos}. Out of the 57 pairs that were true negative in \texttt{LexNET} and false positive in \texttt{Cos}, we judged only one as somehow related (\emph{(death, man)}).

We sampled 25 (from the 184) true positive pairs in \texttt{LexNET+Cos} that were false negatives in \texttt{Cos}, and found that they were all connected via paths in the corpus, suggesting that \texttt{LexNET}'s contribution comes also from the path-based component, rather than only from adding distributional information. 12 of the pairs contained a polysemous term, for which the relation holds in a specific sense (e.g. \emph{(fire, shoot)}). 5 other pairs had a weak relation, e.g. \emph{(compact, car)}. While a \emph{car} can be \emph{compact}, non of these words is one of the most related words to the other.\footnote{\emph{car} is mostly related to \emph{driver}, \emph{cars}, and \emph{race}, and \emph{compact} to \emph{compactness} and \emph{locally}.} As noted by \newcite{shwartz2016roles}, these are cases in which distributional methods may fail to identify the relation between the words, while even a single meaningful path connecting $x$ and $y$ can capture the relation between them.

\paragraph{Subtask 2: Semantic Relation Classification}

We note that the overall results on this task are low, in contrast to the success of several methods on common datasets \cite{baroni2012entailment,weeds2014learning,roller2014inclusive,shwartz2016roles}. One possible explanation is the stricter and more informative evaluation, that considers the \textsc{random} class as noise, discarding it from the $F_1$ average.\footnote{When the random class is included in the averaged $F_1$ score, the results are: P = 0.780, R = 0.786, $F_1$ = 0.781.} Additionally, the dataset is lexically split, disabling lexical memorization \cite{levy2015supervised}. However, the strict evaluation spots a light on the difficulty of this task, which was somewhat obfuscated by the strong results published so far, but might have been obtained thanks to dataset and evaluation peculiarities \cite{levy2015supervised,santus2016nine,shwartz2016roles}.

Figure~\ref{fig:confusion_matrix} displays \texttt{LexNET}'s per relation $F_1$ scores on the test set, with the corresponding confusion matrix. 
While the $F_1$ scores of individual classes are relatively low, the confusion matrix shows that pairs were always classified to the correct relation more than to any other class. 
A common error comes from subtask 1's model: while most unrelated pairs were classified as unrelated, many related pairs were also classified as unrelated. This may be solved in the future by learning the two subtasks jointly rather than applying a pipeline.

Among the other relations, the performance on synonyms was the worst. The path-based component is weak in recognizing synonyms, which do not tend to co-occur. 
The distributional information causes confusion between synonyms and antonyms, since both tend to occur in the same contexts. 
Moreover, synonyms were also sometimes mistaken with hypernyms, as the difference between the two relations is often subtle \cite{shwartz2016improving}.

\begin{figure*}[!t]
\center
\small
\begin{tabular}{lcr}
\multicolumn{1}{l}{
\begin{tabularx}{0.326\textwidth}{ccc|c|c|c|c|}
& & \multicolumn{4}{c}{predictions} \\
\hhline{~~-----}
& & \multicolumn{1}{|c|}{ANT} & RANDOM & HYPER & PART\_OF & SYN \\
\hhline{~------}
\multirow{5}{*}{\rotatebox{90}{\centering gold}} & \multicolumn{1}{|c|}{ANT} & \cellcolor{brinkpink!40.277778}{40.28}  &  \cellcolor{brinkpink!30.277778}{30.28}  &  \cellcolor{brinkpink!5.555556}{5.56}  &  \cellcolor{brinkpink!5.833333}{5.83}  &  \cellcolor{brinkpink!18.055556}{18.06} \\
\hhline{~------}
& \multicolumn{1}{|c|}{RANDOM} &  \cellcolor{brinkpink!2.353710}{2.35}  &  \cellcolor{brinkpink!93.069631}{93.07}  &  \cellcolor{brinkpink!1.307617}{1.31}  &  \cellcolor{brinkpink!1.372998}{1.37}  &  \cellcolor{brinkpink!1.896044}{1.90} \\
\hhline{~------}
& \multicolumn{1}{|c|}{HYPER} &  \cellcolor{brinkpink!10.209424}{10.21}  &  \cellcolor{brinkpink!22.774869}{22.77}  &  \cellcolor{brinkpink!45.811518}{45.81}  &  \cellcolor{brinkpink!6.020942}{6.02}  &  \cellcolor{brinkpink!15.183246}{15.18} \\
\hhline{~------}
& \multicolumn{1}{|c|}{PART\_OF} &  \cellcolor{brinkpink!2.232143}{2.23}  &  \cellcolor{brinkpink!37.053571}{37.05}  &  \cellcolor{brinkpink!6.696429}{6.70}  &  \cellcolor{brinkpink!47.767857}{47.77}  &  \cellcolor{brinkpink!6.250000}{6.25} \\
\hhline{~------}
& \multicolumn{1}{|c|}{SYN} &  \cellcolor{brinkpink!25.957447}{25.96}  &  \cellcolor{brinkpink!20.425532}{20.43}  &  \cellcolor{brinkpink!14.468085}{14.47}  &  \cellcolor{brinkpink!7.234043}{7.23}  &  \cellcolor{brinkpink!31.914894}{31.91} \\
\hhline{~------}
\end{tabularx}} & ~~~~~~~~~~~~~~~~~~~~~~~~~~~~~~~~~~~~~~~~~~~~~~~~~~~ & 
\multicolumn{1}{r}{
\begin{tabular}{ c c | c | c | }
    \\\hline
		\multicolumn{1}{|c|}{\textbf{relation}} & \textbf{P} & \textbf{R} & \boldmath $F_1$ \\ \hline
		\multicolumn{1}{|c|}{ANT} & 0.450 & 0.403 & 0.425 \\ \hline
		\multicolumn{1}{|c|}{RANDOM} & 0.897 & 0.931 & 0.914 \\ \hline
		\multicolumn{1}{|c|}{HYPER} & 0.616 & 0.458 & 0.526 \\ \hline
		\multicolumn{1}{|c|}{PART\_OF} & 0.510 & 0.478 & 0.493 \\ \hline
		\multicolumn{1}{|c|}{SYN} & 0.278 & 0.319 & 0.297 \\ \hline
	\end{tabular}}
\end{tabular}
	\caption{Left: confusion matrix of \texttt{LexNET}'s predictions to the subtask 2 test set. Rows indicate gold labels and columns indicate predictions. The value in $[i, j]$ is the percentage of pairs classified to relation $j$ of those labeled $i$. Right: Per-relation $F_1$ scores of \texttt{LexNET}'s predictions to the test set of subtask 2.}
	\label{fig:confusion_matrix}
	\vspace*{-10pt}
\end{figure*}

\section{Conclusion}

We have presented our submission to the CogALex 2016 shared task on corpus-based identification of semantic relations. 
The submission is based on \texttt{LexNET} \cite{shwartz2016roles}, an integrated path-based and distributional method for semantic relation classification. \texttt{LexNET} was the best-performing system on subtask 2, demonstrating the utility of integrating the complementary path-based and distributional information sources in recognizing semantic relatedness.

We have shown that subtask 1 (word relatedness) reaches reasonable performance with cosine similarity, and is slightly improved when combined with \texttt{LexNET}, especially when the relation between the words is non-prototypical. The performance on subtask 2, however, was relatively low for all systems that participated in the shared task, including \texttt{LexNET}. This demonstrates the difficulty of the semantic relation classification task, and emphasizes the need to develop improved methods for this task, possibly using additional sources of information.

\section*{Acknowledgments}
This work was partially supported by an Intel ICRI-CI grant, the Israel Science Foundation grant 880/12, and the German Research Foundation through the German-Israeli Project Cooperation (DIP, grant DA 1600/1-1).

\bibliography{cogalex}
\bibliographystyle{acl}

\end{document}